\documentclass[fleqn]{tlp}

\usepackage{amsmath,amssymb}
\usepackage{dsfont}
\usepackage{times,helvet,courier}
\usepackage{tikz}\usetikzlibrary{matrix,arrows,calc}

\title[Network Reconstruction in ASP]{Automatic Network Reconstruction using ASP}
\author[Max Ostrowski et al.]{%
  Markus Durzinsky
  and
  Wolfgang Marwan
  \\
  Magdeburg Centre for Systems Biology, Universit\"at Magdeburg
  \and
  Max Ostrowski
  and
  Torsten Schaub\thanks{Affiliated with Simon Fraser University, Canada, and Griffith University, Australia.}
  \\
  Universit\"at Potsdam
  \and
  Annegret Wagler
  \\
  Universit\'e Blaise Pascal, Clermont-Ferrand
}

\submitted{[TBA]}
\revised{[TBA]}
\accepted{[TBA]}

\newcommand{\SpecieS}{\ensuremath{S}}
\newcommand{\ResponseS}{\ensuremath{E_R}}
\newcommand{\PerturbationS}{\ensuremath{E_P}}

\newcommand{\Network}{\ensuremath{\mathcal{R}}}
\newcommand{\PO}{\ensuremath{\prec}}
\newcommand{\Instance}{\ensuremath{\mathcal{I}}}
\newcommand{\Bound}{\ensuremath{\mathcal{B}}}
\newcommand{\ReactionVector}[2]{\ensuremath{\gamma_{#1}{(#2)}}}

\newcommand{\SPECIESFONT}[1]{{\ensuremath{\mathit{#1}}}} % SPECIES in FIGURES

\newcommand{\sysfont}[1]{\textit{#1}}

\newcommand{\gringo}{\sysfont{gringo}}
\newcommand{\lparse}{\sysfont{lparse}}
\newcommand{\clasp}{\sysfont{clasp}}

\newcommand{\head}[1]{\ensuremath{\mathit{head}(#1)}} % {\ensuremath{H(#1)}}
\newcommand{\body}[1]{\ensuremath{\mathit{body}(#1)}} % {\ensuremath{B(#1)}}
\newcommand{\summ}[0]{\ensuremath{\mathtt{\#sum}}}
\newcommand{\countt}[0]{\ensuremath{\mathtt{\#count}}}

\newcommand{\naf}{\ensuremath{{\sim}}}
\newcommand{\fact}[1]{\ensuremath{#1.}}
\newcommand{\impl}[2]{\begin{split} #1  {} \ensuremath{\leftarrow}\ & #2.\end{split}}

\newcommand{\progtest}[2]{\begin{center}\begin{tabular}{#1}#2\end{tabular}\end{center}}
\newcommand{\prog}[1]{\begin{align}{#1}\end{align}}

\newcommand{\atomfont}[1]{\ensuremath{\mathit{#1}}}

\newcommand{\speciesn}{\atomfont{species}}
\newcommand{\capacityn}{\atomfont{capacity}}
\newcommand{\staten}{\atomfont{state}}
\newcommand{\edgen}{\atomfont{edge}}
\newcommand{\perturbationn}{\edgen}
\newcommand{\responsen}{\edgen}
\newcommand{\terminalstaten}{\atomfont{terminalState}}
\newcommand{\amountn}{\atomfont{value}}

\newcommand{\decreaseonen}{\atomfont{decrease}}

\newcommand{\pathn}{\atomfont{path}}

\newcommand{\maxaddsn}{\atomfont{maxAdds}}

\newcommand{\addspeciesn}{\atomfont{addSpecies}}

\newcommand{\addchangen}{\atomfont{addChange}}

\newcommand{\maxreactionsn}{\atomfont{maxReactions}}

\newcommand{\reactionn}{\atomfont{r}}
\newcommand{\napplicableterminaln}{\atomfont{disabled}}

\newcommand{\intervaluen}{\atomfont{interValue}}
\newcommand{\applicablen}{\atomfont{enabled}}
\newcommand{\napplicablen}{\atomfont{disabled}}
\newcommand{\slowern}{\atomfont{slower}}

\newcommand{\fastestn}{\atomfont{fastest}}

\newcommand{\unequaln}{\atomfont{neq}}
\newcommand{\associatedTerminaln}{\atomfont{t}}

\newcommand{\stepn}{\atomfont{step}}
\newcommand{\lengthn}{\atomfont{length}}

\newcommand{\species}[1]{\ensuremath{\speciesn(#1)}}
\newcommand{\capacity}[2]{\ensuremath{\capacityn(#1,#2)}}
\newcommand{\state}[1]{\ensuremath{\staten(#1)}}
\newcommand{\perturbation}[2]{\ensuremath{\perturbationn(p,#1,#2)}}
\newcommand{\edge}[3]{\ensuremath{\edgen(#1,#2,#3)}}
\newcommand{\terminalstate}[1]{\ensuremath{\terminalstaten(#1)}}
\newcommand{\amount}[3]{\ensuremath{\amountn(#1,#2,#3)}}

\newcommand{\decreaseone}[1]{\ensuremath{\decreaseonen(#1)}}

\newcommand{\path}[2]{\ensuremath{\pathn(#1,#2)}}

\newcommand{\maxadds}[1]{\ensuremath{\maxaddsn(#1)}}

\newcommand{\addspecies}[1]{\ensuremath{\addspeciesn(#1)}}

\newcommand{\maxreactions}[1]{\ensuremath{\maxreactionsn(#1)}}

\newcommand{\reactionone}[1]{\ensuremath{\reactionn(#1)}}
\newcommand{\reactionthree}[3]{\ensuremath{\reactionn(#1,#2,#3)}}
\newcommand{\responsethree}[3]{\ensuremath{\lengthn(#1,#2,#3)}}
\newcommand{\response}[2]{\ensuremath{\responsen(r,#1,#2)}}

\newcommand{\napplicableterminal}[2]{\ensuremath{\napplicableterminaln(#1,#2)}}

\newcommand{\intervalue}[4]{\ensuremath{\intervaluen(#1,#2,#3,#4)}}
\newcommand{\applicable}[3]{\ensuremath{\applicablen(#1,#2,#3)}}
\newcommand{\napplicable}[3]{\ensuremath{\napplicablen(#1,#2,#3)}}
\newcommand{\slower}[2]{\ensuremath{\slowern(#1,#2)}}

\newcommand{\addchangeone}[1]{\ensuremath{\addchangen(#1)}}

\newcommand{\fastest}[3]{\ensuremath{\fastestn(#1,#2,#3)}}

\newcommand{\unequal}[2]{\ensuremath{\unequaln(#1,#2)}}
\newcommand{\associatedTerminal}[2]{\ensuremath{\associatedTerminaln(#1,#2)}}

\newcommand{\step}[2]{\ensuremath{\stepn(#1,#2)}}

\newtheorem{definition}{Definition} % [section]
\newtheorem{proposition}{Proposition} % [section]

\tikzset{
  species/.style={
	 circle,
	 minimum size=5mm,
	 inner sep=0mm,
	  draw=black,
	 font=\itshape
	 },
	reaction/.style={
	 rectangle,
	 minimum size=5mm,
	 very thick,
	 draw=black,
	 font=\ttfamily}
}

\begin{document}

\maketitle

\begin{abstract}
  Building biological models by inferring functional dependencies from
  experimental data is an important issue in Molecular Biology. 
  To relieve the biologist from this traditionally manual process,
  various approaches have been proposed to increase the degree of automation.
  However, available approaches often 
  yield a single model only,
  rely on specific assumptions,
  and/or
  use dedicated, heuristic algorithms that are intolerant to changing
  circumstances or requirements
  in the view of the rapid progress made in Biotechnology. 
  Our aim is to provide a declarative solution to the problem by appeal to
  Answer Set Programming (ASP) overcoming these difficulties. 
  We build upon an existing approach to Automatic Network
  Reconstruction proposed by part of the authors.
  This approach has firm mathematical foundations and is well suited for ASP
  due to its combinatorial flavor providing a characterization of all models
  explaining a set of experiments.
  The usage of ASP has several benefits over the existing heuristic algorithms.
  First, it is declarative and thus transparent for biological experts.
  Second, it is elaboration tolerant and thus allows for an easy exploration and
  incorporation of biological constraints.
  Third, it allows for exploring the entire space of possible models.
  Finally, our approach offers an excellent performance,
  matching existing, special-purpose systems.
\end{abstract}

%%% Local Variables: 
%%% mode: latex
%%% TeX-master: "paper"
%%% End: 

\section{Introduction}\label{sec:introduction}

The creation of biological models by inferring functional dependencies from
experimental data is a key issue in molecular biology.
A common approach is to construct descriptive models from series of
experiments.
This (manual) process usually starts from a model defined using existing
biological knowledge which is then gradually refined by appeal to data gathered
in subsequent experiments.
A model obtained this way is however merely consistent with the gathered
experimental data,
and, besides simulation,
no true indication can be given how well the resulting model captures the
biological system.
For instance, it is unclear whether the obtained model is one among many or few
alternative models.
Moreover, it is of great interest to know the difference among alternative
models in order to design new experiments for further discriminating the best
fitting model.

This problem is addressed in the area of \emph{Automatic Network Reconstruction} (ANR)%
~\cite{gifjaa01a,yeteco02a,lifuso98a,relian02a}.
However, the available approaches often 
yield a single model only,
rely on specific assumptions,
and/or
use dedicated, heuristic algorithms for constructing a model from experimental data.
Moreover, all these approaches are intolerant to changing circumstances or requirements
in the view of the rapid progress made in Biotechnology.
Unlike this,
we provide a declarative solution to the problem by appeal to Answer Set
Programming (ASP;~\cite{baral02a}).
To this end, we build upon the approach to ANR proposed in~\cite{preprint,marwagwei08a}.
This approach has firm mathematical foundations and is well suited for ASP
due to its combinatorial flavor providing a characterization of all models
explaining a set of experiments.
The usage of ASP has several benefits over the existing heuristic algorithms.
First, it is declarative and thus transparent for biological experts.
Second, it is elaboration tolerant and thus allows for an easy exploration and
incorporation of biological constraints.
Third, it allows for exploring the entire space of possible models.
Finally, our approach offers an excellent performance,
matching existing, special-purpose systems.

The next section gives a formal introduction to ANR, 
as provided in \cite{preprint,marwagwei08a},
followed by a brief introduction to ASP in Section~\ref{sec:asp}.
Section~\ref{sec:approach} is dedicated to our solution to ANR in ASP.
We empirically evaluate our approach in Section~\ref{sec:experiments}
and conclude with a discussion and a summary in Section~\ref{sec:discussion} and~\ref{sec:summary}.

%%% Local Variables: 
%%% mode: latex
%%% TeX-master: "paper"
%%% End: 

\section{Automatic Network Reconstruction}\label{sec:problem}

Automatic Network Reconstruction aims at constructing all models explaining a
set of (perturbation) experiments 
reflecting a certain biological process.
Our approach starts from experimental time-series data and generates all
interaction networks that account for the observed mass or signal flow.  
We briefly describe the steps of this approach proposed in \cite{preprint,DWW_LNCS,marwagwei08a}.

%%%We represent a collection \SpecieS\ of $n$ observable species as a vector $(s_1,\dots,s_n)$ along with a corresponding vector $(D_1,\dots,D_n)$ of associated domains over $\mathds{N}_0$. Accordingly, species $s_i$ is assigned a value from domain $D_i$ for $1\leq i\leq n$.
%
We represent a collection \SpecieS\ of $n$ observable species as a vector $(s_1,\dots,s_n)$ being considered to be crucial for describing the studied biological phenomenon,
along with a corresponding vector $(D_1,\dots,D_n)$ of associated capacities over $\mathds{N}_0$.
Accordingly, species $s_i$ is assigned a value from capacitiy $D_i$ for $1\leq i\leq n$.
A \emph{state} $x$ of species $(s_1,\dots,s_n)$ is a vector
$(x_1,\dots,x_n)$ such that $x_i\in D_i$ for $1\leq i\leq n$.
Thus, 
$x_i$ provides the value of species $s_i$ in state $x$ for $1 \leq i \leq n$.
Note that our concept of a state is only partial because it is confined to the
observable species in \SpecieS.
In what follows, we leave the set \SpecieS\ of species implicit whenever clear
from the context.

A (perturbation) \emph{experiment} $
\mathcal{E}(x^0)=
(x^0;x^1,\dots,x^k)$ over \SpecieS\ is a sequence of states 
%in ${\cal X}$
reflecting the time-dependent response $(x^1,\dots,x^k)$ of a biological system to
a (specific) perturbation of the system in state $x^0$.
We associate with each response state $x^i 
\in \mathcal{E}(x^0)
$ its \emph{terminal state} $x^k 
\in \mathcal{E}(x^0)
$ and define $t(x^i)=x^k$ for all $1 \leq i < k$.
%
%For instance,
%\(
%(x_0;x_1,x_2,x_3,x_4)
%\)
%is an experiment % over species $\{\SPECIESFONT{fr}, \SPECIESFONT{r} and \SPECIESFONT{spo}\}$
%with terminal state $x_4$.

\begin{figure}
\centering
\begin{tikzpicture}[remember picture]
	\matrix (experiment) [row sep=0.5cm,column sep=0.8cm]
	{
		\node[name=x0] {$\begin{array}{@{}c@{}}x_0\\ \left( \begin{array}{@{\;}c@{\;}}0\\0\\0\end{array}\right)\end{array} $};
		&
		\node[name=x1] {$\begin{array}{@{}c@{}}x_1\\ \left( \begin{array}{@{\;}c@{\;}}1\\0\\0\end{array}\right)\end{array} $};
		&
		\node[name=x2] {$\begin{array}{@{}c@{}}x_2\\ \left( \begin{array}{@{\;}c@{\;}}0\\0\\0\end{array}\right)\end{array} $};
		&
		\node[name=x3] {$\begin{array}{@{}c@{}}x_3\\ \left( \begin{array}{@{\;}c@{\;}}0\\0\\0\end{array}\right)\end{array} $};
		&
		&
		\node[name=x4] {$\begin{array}{@{}c@{}}x_4\\ \left( \begin{array}{@{\;}c@{\;}}0\\0\\1\end{array}\right)\end{array} $};
		\\
		&
		&
		\node[name=x5] {$\begin{array}{@{}c@{}}x_5\\ \left( \begin{array}{@{\;}c@{\;}}0\\1\\0\end{array}\right)\end{array} $};
		&
		\node[name=x6] {$\begin{array}{@{}c@{}}x_6\\ \left( \begin{array}{@{\;}c@{\;}}0\\1\\0\end{array}\right)\end{array} $};
		&
		\node[name=x7] {$\begin{array}{@{}c@{}}x_7\\ \left( \begin{array}{@{\;}c@{\;}}0\\0\\0\end{array}\right)\end{array} $};
		&
		\node[name=x8] {$\begin{array}{@{}c@{}}x_8\\ \left( \begin{array}{@{\;}c@{\;}}0\\0\\1\end{array}\right)\end{array} $};
		&
		\node[name=empty] {};
		\\
	};
	\path[->] (x0.east) edge[->, dashed] (x1.west);
	\path[->] (x1.east) edge[->] (x2.west);
	\path[->] (x2.east) edge[->] (x3.west);
	\path[->] (x3.east) edge[->] (x4.west);
	\path[->] (x2.south) edge[->,dashed] (x5.north);
	\path[->] (x3.south) edge[->,dashed] (x6.north);
	\path[->] (x6.east) edge[->] (x7.west);
	\path[->] (x7.east) edge[->] (x8.west);
	\path[->] (x5.west) edge[->,bend left=30] (x0.south);
	\path[->] (x7.east) edge[->] (x8.west);
	\path[->] (x3.east) edge[->] (x4.west);
%	\path<6->[->] (x1.east) edge[->] node [above]{\color<6>{red}$r_1$} (x2.west);
%	\path<6->[->] (x2.east) edge[->] node [above]{\color<6>{red}$r_2$} (x3.west);
%	\path<6->[->] (x3.east) edge[->] node [above]{\color<6>{red}$r_3+r_1$} (x4.west);
%	\path<6->[->] (x6.east) edge[->] node [above]{\color<6>{red}$r_4$} (x7.west);
%	\path<6->[->] (x7.east) edge[->] node [above]{\color<6>{red}$r_5$} (x8.west);
%	\path<6->[->] (x5.west) edge[->,bend left=30] node [above, yshift=5mm]{\color<6>{red}$r_2 + r_4$}(x0.south);
%	\path<8->[->] (x8.east) edge[->] node [above]{\color<8>{red}$r_?i$}(empty.west);
\end{tikzpicture}
%
%%% Local Variables: 
%%% mode: latex
%%% TeX-master: "../paper"
%%% End: 
\vspace{-10pt}
\caption{\label{figexpgraph}An Experiment Graph $G(\mathcal{E})$.}
\end{figure}
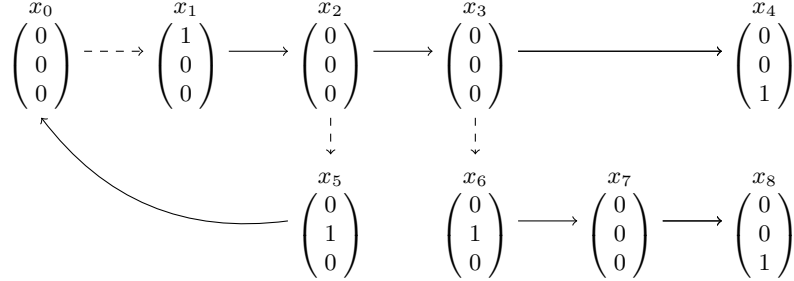

Typically, several experiments $\mathcal{E}(x^0)$ starting from different initial
states $x^0$ are necessary to describe a biological phenomenon.  
We encode a set $\mathcal{E}$ of different experiments in terms of
an \emph{experiment graph} $
G(\mathcal{E})=
(X,\PerturbationS\cup\ResponseS)$ over \SpecieS, which is a directed graph such that 
$X$ is the multi-set of states in $\mathcal{E}$,
and \PerturbationS\ and \ResponseS\ are disjoint sets of perturbation and response edges, respectively.
That is,
for each $
\mathcal{E}(x^0) =
(x^0;x^1,\dots,x^k) \in 
\mathcal{E}$, %\ExperimentS$,
we have $(x^0,x^1)\in \PerturbationS$ and $(x^i,x^{i+1})\in \ResponseS$ for $1 \leq i < k$. 
For illustration,
consider Figure~\ref{figexpgraph} showing an experiment graph
over species $\{\SPECIESFONT{fr}, \SPECIESFONT{r}, \SPECIESFONT{spo}\}$,
encoding three experiments $\mathcal{E}(x^0)=(x^0;x^1,\dots,x^4)$, $\mathcal{E}(x^2)=(x^2;x^5,x^0)$, and $\mathcal{E}(x^3)=(x^3;x^6,\dots,x^8)$. 
%sharing domain $\{0,1\}$.
The entries in each state vector give the respective values of each species; 
continuous arrows represent response edges, dashed ones give perturbation edges. 
% ------------------------------------------------------------
% ------------------------------------------------------------

An experiment graph $G(\mathcal{E})=(X,\PerturbationS\cup\ResponseS)$ is \emph{valid} if %%%we have 
\begin{enumerate}
\renewcommand{\theenumi}{\textbf{\Roman{enumi}}}
\item
every state $x \in X$ has at most one outgoing arc in \ResponseS,
\item
$x = x'$ implies $t(x) = t(x')$ for all $x,x' \in X$ and 
\item $(x'-x) \not\in \mathds{N}^n$ holds 
%%%$x'_i - x_i < 0$ for some $0 \leq i \leq n$ 
for all $(x,x') \in E_R$, %%%% and that
\end{enumerate}
Condition~\textbf{I} stipulates that an experiment graph is deterministic,
while \textbf{II} requires that no equal%
\footnote{Recall that $X$ is a multi-set; two states are equal if their vector of species is equal} 
states lead to different terminal states.
\textbf{III} demands that there must be at least one species that decreases
between two consecutive response states.
In fact, the experiment graph in Figure~\ref{figexpgraph} violates two validity conditions: 
\textbf{II} is violated through states $x^5$ and $x^6$, as these states are equal
but lead to differing terminal states $x^0$ and $x^8$, respectively.
\textbf{III} is violated by response edge $(x^2,x^3)$.

For the reconstruction, we use the paradigm that system states can be changed by applying reactions. 
A \emph{reaction} over $n$ species is described by a vector $r\in\mathds{Z}^n$, where 
$r_i < 0$ for some $1 \leq i \leq n$.
So, a reaction must have at least one negative entry to consume at least one species.
A reaction $r$ is \emph{enabled} in a state $x$ over $n$ species with capacities $(D_1,\dots,D_n)$,
if we have $x_i + r_i \in D_i$ for all $1\leq i\leq n$, i.e.~if neither nonnegativity nor capacity constraints are violated.
For instance,
reaction $r=(0,-1,0)$ is enabled in $x^6$ because $x^6+r=(0,0,0)$ belongs to
the species' capacity. 

Given an experiment graph $(X,\PerturbationS\cup\ResponseS)$ and 
a response edge $(x,x')\in \ResponseS$, we say that this response is \emph{realized} by 
a sequence $\sigma((x,x')) = (r^1, \dots, r^l)$ of reactions, if
\begin{enumerate}
\renewcommand{\theenumi}{\textbf{\Roman{enumi}}}
\setcounter{enumi}{3}
\item $y^i + r^i = y^{i+1}$ for all $1 \leq i \leq l$, and
\item $(y^1,y^2,\dots,y^{l+1})$ is a sequence of states such that $x=y^1$
  and $x'=y^{l+1}$,
\item $r^i_k \cdot r^j_k \geq 0$ for all $1 \leq i,j \leq l$ and all $0 \leq k \leq n$.
\end{enumerate}
All reactions subsequently applied to state $x$ fulfill the response edge and
ultimately lead to the consecutively observed state $x'$ in \ResponseS.
For example, the reaction $r=(0,-1,0)$ constitutes a singleton
sequence $\sigma((x^6,x^7))$ as $x^6+r=x^7$ realizes $(x^6,x^7)$.
Note that \textbf{VI} stipulates that all reactions in such a sequence
must be \emph{monotone};%
\footnote{This is a significant constraint on the quality of time series data.
  The response of the system must be measured with sufficient time resolution,
  such that oscillation between measurements can be excluded.}
at microscopic level, a species cannot be produced and consumed (or vice
versa) by two reactions, see~\cite{DWW_LNCS} for details.

To also account for the experimentally observed mass or signal flow,
\cite{marwagwei08a} propose to use a partial order on the set of reactions to
reflect their relative rates. % reflecting their relative speed,
A sequence $(r^1, \dots, r^l)$ of reactions is said to respect such a partial order \PO,
if $r^i$ is the unique \PO-minimal reaction enabled in an (intermediate) state
$y^i$ for each $1 \leq i < l$.
Note that the reaction order \PO\ must be sufficiently strong to guarantee a
unique fastest reaction at each step.
This implies for each state to have a unique successor state,
ensuring the system's determinism.

Following~\cite{marwagwei08a},
a \emph{regulatory structure} $(\Network,\PO)$ over species $S$
consists of a set of reactions \Network\ and a partial order~\PO\ among them.%
\footnote{%
  \Network\ is also referred to as a \emph{network} because such reaction sets
  are easily converted to \emph{Petri nets}, as done in~\cite{marwagwei08a}.
  This is however beyond the scope of this paper.}
A regulatory structure $(\Network,\PO)$ is \emph{conformal} with a valid experiment graph
$(X,\PerturbationS\cup\ResponseS)$,
if 
\begin{enumerate}
\renewcommand{\theenumi}{\textbf{\Roman{enumi}}}
\setcounter{enumi}{6}
\item for all $r\in \Network$, $r$ is not enabled in any terminal state of $X$,
\item for all $e\in \ResponseS$, 
  there is a \PO-respecting realizing sequence%
\footnote{We slightly abuse notation, and take $\sigma(e) \subseteq \Network$ to
  mean that each element of $\sigma(e)$ is also in $\Network$.}
	$\sigma(e) \subseteq \Network$, and
\item there exists no $r\in \Network$ where $r$ is not an element of some $\sigma(e)$.
\end{enumerate}

As defined in~\cite{preprint},
the \emph{Network Reconstruction Problem}
for a valid experiment graph
consists in finding all regulatory structures conformal with the graph.

An invalid experiment graph can be recovered by adding new, artificial species
to \SpecieS.%
\footnote{%
This allows for differentiating seemingly equal yet different states,
enabling new reactions by decreasing additional species,
or
avoiding reactions in terminal states.}
Given an (invalid) experiment graph $(X,\PerturbationS\cup\ResponseS)$,
an extension $(X',\PerturbationS\cup\ResponseS)$ with $a$ species 
is obtained by replacing each state $(x_1,\dots,x_n)$ in $X$
with $(x_1,\dots,x_n,x_{n+1},\dots,x_{n+a})$
such that $x_{n+i}\in \{0,1\}$ for $1\leq i\leq a$;
all other capacities and edges are left intact.
Note that an experiment graph has $2^a$ extensions.

An extension $(X',\PerturbationS\cup\ResponseS)$ of an experiment graph with $a$
species is \emph{valid},
if
\begin{enumerate}
\renewcommand{\theenumi}{\textbf{\Roman{enumi}}}
\setcounter{enumi}{9}
\item $(X',\PerturbationS\cup\ResponseS)$ is a valid experiment graph and
\item $x_{n+i} = x'_{n+i}$ for each $(x,x')\in \PerturbationS$ and $1\leq i\leq a$.
\end{enumerate}
The latter condition stipulates that additional species are not direct targets of
experimental perturbations, but they certainly respond in successive states.
Similarly, we want to reduce the changes of additional species in response edges:
%\REPc{An \emph{additional change} occurs in a response edge$ (x,x')\in \ResponseS$,
%if $x_{n+i} \neq x'_{n+i}$ for some $1 \leq i \leq a$.}%
A response edge $(x,x')\in \ResponseS$ is subject to an \emph{additional change},
if $x_{n+j} \neq x'_{n+j}$ for some $1 \leq j \leq a$.

At last, given an invalid experiment graph, the \emph{Network Reconstruction Problem}
consists in solving the NRP for all valid extensions of that graph, first, adding a minimum
number of additional species and, second, comprising a minimum number of additional changes.
For brevity, such extensions are called \emph{minimal valid extensions}.

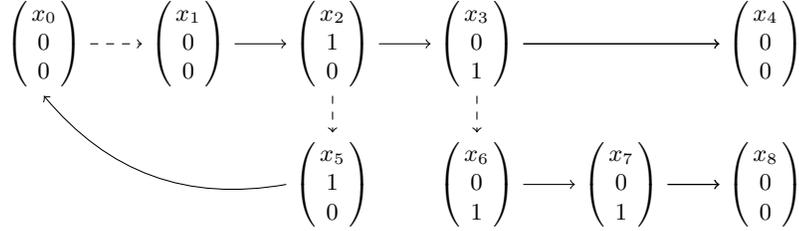
\begin{figure}
\centering
\begin{tikzpicture}[remember picture]
	\matrix (experiment) [row sep=0.5cm,column sep=0.7cm]
	{
		\node[name=x0] {$\left( \begin{array}{@{\,}c@{\,}}x_0\\0\\0\end{array}\right)$};
		&
		\node[name=x1] {$\left( \begin{array}{@{\,}c@{\,}}x_1\\0\\0\end{array}\right)$};
		&
		\node[name=x2] {$\left( \begin{array}{@{\,}c@{\,}}x_2\\1\\0\end{array}\right)$};
		&
		\node[name=x3] {$\left( \begin{array}{@{\,}c@{\,}}x_3\\0\\1\end{array}\right)$};
		&
		&
		\node[name=x4] {$\left( \begin{array}{@{\,}c@{\,}}x_4\\0\\0\end{array}\right)$};
		\\
		&
		&
		\node[name=x5] {$\left( \begin{array}{@{\,}c@{\,}}x_5\\1\\0\end{array}\right)$};
		&
		\node[name=x6] {$\left( \begin{array}{@{\,}c@{\,}}x_6\\0\\1\end{array}\right)$};
		&
		\node[name=x7] {$\left( \begin{array}{@{\,}c@{\,}}x_7\\0\\1\end{array}\right)$};
		&
		\node[name=x8] {$\left( \begin{array}{@{\,}c@{\,}}x_8\\0\\0\end{array}\right)$};
		&
		\node[name=empty] {};
		\\
	};
	\path[->] (x0.east) edge[->, dashed] (x1.west);
	\path[->] (x1.east) edge[->] (x2.west);
	\path[->] (x2.east) edge[->] (x3.west);
	\path[->] (x3.east) edge[->] (x4.west);
	\path[->] (x2.south) edge[->,dashed] (x5.north);
	\path[->] (x3.south) edge[->,dashed] (x6.north);
	\path[->] (x6.east) edge[->] (x7.west);
	\path[->] (x7.east) edge[->] (x8.west);
	\path[->] (x5.west) edge[->,bend left=30] (x0.south);
	\path[->] (x7.east) edge[->] (x8.west);
	\path[->] (x3.east) edge[->] (x4.west);
%	\path<6->[->] (x1.east) edge[->] node [above]{\color<6>{red}$r_1$} (x2.west);
%	\path<6->[->] (x2.east) edge[->] node [above]{\color<6>{red}$r_2$} (x3.west);
%	\path<6->[->] (x3.east) edge[->] node [above]{\color<6>{red}$r_3+r_1$} (x4.west);
%	\path<6->[->] (x6.east) edge[->] node [above]{\color<6>{red}$r_4$} (x7.west);
%	\path<6->[->] (x7.east) edge[->] node [above]{\color<6>{red}$r_5$} (x8.west);
%	\path<6->[->] (x5.west) edge[->,bend left=30] node [above, yshift=5mm]{\color<6>{red}$r_2 + r_4$}(x0.south);
%	\path<8->[->] (x8.east) edge[->] node [above]{\color<8>{red}$r_?i$}(empty.west);
\end{tikzpicture}
%
%%% Local Variables: 
%%% mode: latex
%%% TeX-master: "../paper"
%%% End: 
\vspace{-10pt}
\caption{\label{fig:ext:one}
First Extension 
$G(\mathcal{E}_1)$
of the Experiment Graph 
$G(\mathcal{E})$
in Figure~\ref{figexpgraph}.}
\end{figure}

Figure~\ref{fig:ext:one}
and~\ref{fig:ext:two} show the two\footnote{Actually, there are four
  extensions with symmetric behavior on the additional species.} valid
extensions of the invalid experiment graph in Figure~\ref{figexpgraph}.
The nodes in the figures are the vectors from Figure~\ref{figexpgraph}
extended by the two additional species.
%
% ------------------------------------------------------------
\begin{figure}
\centering
\begin{tikzpicture}[remember picture]
	\matrix (experiment) [row sep=0.5cm,column sep=0.7cm]
	{
		\node[name=x0] {$\left( \begin{array}{@{\,}c@{\,}}x_0\\0\\0\end{array}\right)$};
		&
		\node[name=x1] {$\left( \begin{array}{@{\,}c@{\,}}x_1\\0\\0\end{array}\right)$};
		&
		\node[name=x2] {$\left( \begin{array}{@{\,}c@{\,}}x_2\\1\\1\end{array}\right)$};
		&
		\node[name=x3] {$\left( \begin{array}{@{\,}c@{\,}}x_3\\0\\1\end{array}\right)$};
		&
		&
		\node[name=x4] {$\left( \begin{array}{@{\,}c@{\,}}x_4\\0\\0\end{array}\right)$};
		\\
		&
		&
		\node[name=x5] {$\left( \begin{array}{@{\,}c@{\,}}x_5\\1\\1\end{array}\right)$};
		&
		\node[name=x6] {$\left( \begin{array}{@{\,}c@{\,}}x_6\\0\\1\end{array}\right)$};
		&
		\node[name=x7] {$\left( \begin{array}{@{\,}c@{\,}}x_7\\0\\1\end{array}\right)$};
		&
		\node[name=x8] {$\left( \begin{array}{@{\,}c@{\,}}x_8\\0\\0\end{array}\right)$};
		&
		\node[name=empty] {};
		\\
	};
	\path[->] (x0.east) edge[->, dashed] (x1.west);
	\path[->] (x1.east) edge[->] (x2.west);
	\path[->] (x2.east) edge[->] (x3.west);
	\path[->] (x3.east) edge[->] (x4.west);
	\path[->] (x2.south) edge[->,dashed] (x5.north);
	\path[->] (x3.south) edge[->,dashed] (x6.north);
	\path[->] (x6.east) edge[->] (x7.west);
	\path[->] (x7.east) edge[->] (x8.west);
	\path[->] (x5.west) edge[->,bend left=30] (x0.south);
	\path[->] (x7.east) edge[->] (x8.west);
	\path[->] (x3.east) edge[->] (x4.west);
%	\path<6->[->] (x1.east) edge[->] node [above]{\color<6>{red}$r_1$} (x2.west);
%	\path<6->[->] (x2.east) edge[->] node [above]{\color<6>{red}$r_2$} (x3.west);
%	\path<6->[->] (x3.east) edge[->] node [above]{\color<6>{red}$r_3+r_1$} (x4.west);
%	\path<6->[->] (x6.east) edge[->] node [above]{\color<6>{red}$r_4$} (x7.west);
%	\path<6->[->] (x7.east) edge[->] node [above]{\color<6>{red}$r_5$} (x8.west);
%	\path<6->[->] (x5.west) edge[->,bend left=30] node [above, yshift=5mm]{\color<6>{red}$r_2 + r_4$}(x0.south);
%	\path<8->[->] (x8.east) edge[->] node [above]{\color<8>{red}$r_?i$}(empty.west);
\end{tikzpicture}
%
%%% Local Variables: 
%%% mode: latex
%%% TeX-master: "../paper"
%%% End: 
\vspace{-10pt}
\caption{\label{fig:ext:two}
Second Extension 
$G(\mathcal{E}_2)$
of the Experiment Graph 
$G(\mathcal{E})$
in Figure~\ref{figexpgraph}.}
\end{figure}
% ------------------------------------------------------------
%
Both extensions differ in the values attributed to the two additional species in
state $x_2$ and $x_5$.
The extensions are conformal with the regulatory structures with the networks depicted in
Figure~\ref{fig:sol:one} and~\ref{fig:sol:two}, respectively.
The additional species are referred to as $x$ and $y$.
Reactions are given as boxes, species as circles.
All reaction entries have the capacity $\{-1,0,1\}$.
An arrow from a species to a reaction stands for a $-1$ in the reaction vector,
one from a reaction to a species stands for $+1$.

Accordingly,
the regulatory structure in Figure~\ref{fig:sol:one} over species
$(\SPECIESFONT{fr},\SPECIESFONT{r},\SPECIESFONT{spo},\SPECIESFONT{x},\SPECIESFONT{y})$
comprises the following reactions:
$r^1=(-1,0,0,1,0)$, $r^2=(0,-1,0,0,0)$, $r^3=(0,-1,0,-1,0)$, $r^4=(0,0,0,-1,1)$, and
$r^5=(0,0,1,0,-1)$ and the ordering $\PO = \{(r^4,r^3),(r^3,r^2),(r^5,r^2)\}$.
The regulatory structure in Figure~\ref{fig:sol:two}
comprises the reactions:
$r^1=(-1,0,0,1,1)$, $r^2=(0,-1,0,-1,-1)$, $r^3=(0,-1,0,0,0)$, $r^4=(0,0,1,-1,0)$, and
$r^5=(0,0,0,0,-1)$ and the ordering $\PO = \{(r^4,r^5),(r^5,r^2),(r^4,r^3),(r^3,r^2)\}$.
% ------------------------------------------------------------
\begin{figure}
\centering
\begin{tikzpicture}[remember picture]
			\node[name=fr] at (0,1) [species] {\SPECIESFONT{fr}};
			\node[name=x] at (2,1) [species] {\SPECIESFONT{x}};
			\node[name=y] at (4,1) [species] {\SPECIESFONT{y}};
			\node[name=spo] at (6,1) [species] {\SPECIESFONT{spo}};
			\node[name=r] at (1,0) [species] {\SPECIESFONT{r}};
			\node[name=r1] at (1,1) [reaction] {$r^1$};
			\node[name=r4] at (3,1) [reaction] {$r^4$};
			\node[name=r5] at (5,1) [reaction] {$r^5$};
			\node[name=r2] at (0,0) [reaction] {$r^2$};
			\node[name=r3] at (2,0) [reaction] {$r^3$};

\pgfsetendarrow{\pgfarrowlargepointed{20.5pt}}
		\path[->, thick] (fr) edge[->] (r1);
		\path[->, thick] (r1) edge[->] (x);
		\path[->, thick] (x) edge[->] (r4);
		\path[->, thick] (r4) edge[->] (y);
		\path[->, thick] (y) edge[->] (r5);
		\path[->, thick] (r5) edge[->] (spo);

		\path[->, thick] (x) edge[->] (r3);
		\path[->, thick] (r) edge[->] (r3);
		\path[->, thick] (r) edge[->] (r2);
%		\path[->, thick] (b) edge[->] (r1);
%		\path[->, thick] (r1) edge[->] (d);

%		\path[->,thick] (a) edge[->] (r2);
%		\path[->,thick] (b) edge[->, thick] (r2);
%		\path[->,thick] (c) edge[->] (r2);
%		\path[->,thick] (r2) edge[->] (d);

%		\path[->] (r8.east) edge[->, bend left=30 ] ($(middle.west)-(0,0.4cm)$);

%		\path[->] (r8.east) edge[->, bend left=30 ] ($(middle.west)-(0,0.4cm)$);

%		\path[->] (middle.east) edge[->] (x3.west);
%		\path[->] ($(middle.east)-(0,0.8cm)$) edge[->] (x4.west);
	\end{tikzpicture}
%
%%% Local Variables: 
%%% mode: latex
%%% TeX-master: "../paper"
%%% End: 
\caption{Regulatory Structure conformal with the extended Experiment Graph in Figure~\ref{fig:ext:one}.}
\label{fig:sol:one}
\end{figure}
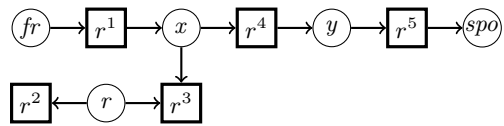
\begin{figure}
\centering
\begin{tikzpicture}[remember picture]
			\node[name=fr] at (0,1) [species] {\SPECIESFONT{fr}};
			\node[name=x] at (2,2) [species] {\SPECIESFONT{x}};
			\node[name=y] at (2,0) [species] {\SPECIESFONT{y}};
			\node[name=spo] at (4,2) [species] {\SPECIESFONT{spo}};
			\node[name=r] at (4,0) [species] {\SPECIESFONT{r}};
			\node[name=r1] at (1,1) [reaction] {$r^1$};
			\node[name=r2] at (1,0) [reaction] {$r^5$};
			\node[name=r3] at (3,2) [reaction] {$r^4$};
			\node[name=r4] at (3,1) [reaction] {$r^2$};
			\node[name=r5] at (3,0) [reaction] {$r^3$};

\pgfsetendarrow{\pgfarrowlargepointed{20.5pt}}
		\path[->, thick] (fr) edge[->] (r1);
		\path[->, thick] (r1) edge[->] (x);
		\path[->, thick] (r1) edge[->] (y);
		\path[->, thick] (y) edge[->] (r2);
		\path[->, thick] (x) edge[->] (r3);
		\path[->, thick] (x) edge[->] (r4);
		\path[->, thick] (y) edge[->] (r4);
		\path[->, thick] (r3) edge[->] (spo);
		\path[->, thick] (r) edge[->] (r4);
		\path[->, thick] (r) edge[->] (r5);

%		\path[->, thick] (x) edge[->] (r5);
		%\path[->, thick] (r) edge[->] (r5);
%		\path[->, thick] (r) edge[->] (r4);
	\end{tikzpicture}
%
%%% Local Variables: 
%%% mode: latex
%%% TeX-master: "../paper"
%%% End:  
\caption{Regulatory Structure conformal with the extended Experiment Graph in Figure~\ref{fig:ext:two}.}
\label{fig:sol:two}
\end{figure}
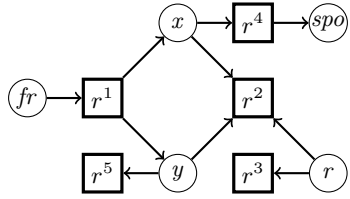
% ------------------------------------------------------------

%%% Local Variables: 
%%% mode: latex
%%% TeX-master: "paper"
%%% End: 

\section{Answer Set Programming}\label{sec:asp}

We rely on the input language of the ASP grounder \gringo~\cite{potasscoManual}
(extending the language of \lparse~\cite{lparseManual})
and introduce only informally the basics of ASP.
A comprehensive, formal introduction to ASP can be found in \cite{baral02a,gelfond08a}.

We consider extended logic programs as introduced in~\cite{siniso02a}.
A \emph{rule}~$r$ is of the following form:
\begin{align*} %\label{eq:rule}%\nonumber
H \leftarrow B_1,\dots,B_m,\naf B_{m+1},\dots,\naf B_n.
\end{align*}
By $\head{r}=H$ and $\body{r}=\{B_1,\dots,B_m,\naf B_{m+1},\dots,\naf B_n\}$,
we denote the \emph{head} and the \emph{body} of~$r$, respectively,
where ``$\naf{}$'' stands for default negation.
The head~$H$ is an atom $a$ belonging to some alphabet~$\mathcal{A}$,
the falsum $\bot$,
or a $\summ$ constraint % of the form
$L\, \summ[\ell_1=w_1,\dots,\ell_k=w_k]\, U$.
In the latter, 
$\ell_i=a_i$ or $\ell_i=\naf a_i$ 
is a \emph{literal} and $w_i$ a \emph{non-negative} integer \emph{weight} for $a_i\in\mathcal{A}$ and $1\leq i\leq k$;
$L$ and $U$ are integers providing a lower and an upper bound.
Either or both of $L$ and $U$ can be omitted,
in which case they are identified with the (trivial) bounds $0$ and~$\infty$, respectively.
Whenever all weights equal one,
the $\summ$ constraint $L\, \summ\{\ell_1=1,\dots,\ell_k=1\}\, U$
becomes a ``$\countt$'' constraint and is simply written as
$L\, \{\ell_1,\dots,\ell_k\}\, U$.
A rule~$r$ such that $\head{r}=\bot$ is an \emph{integrity constraint}.
Each body component $B_i$ is either an atom or a $\summ$ constraint for $1\leq i\leq n$.
If $\body{r}=\emptyset$, $r$ is called a \emph{fact},
and we skip ``$\leftarrow$'' when writing facts below.
We adhere to the definition of answer sets provided in~\cite{siniso02a},
which applies to logic programs containing extended constructs ($\summ$
constraints) under ``choice semantics''.

In addition to rules,
a logic program can contain $\# minimize$ statements of the form
\begin{equation*}
\# minimize \ \{\;\ell_1=w_1,\dots,\ell_k=w_k\;\}.
\end{equation*}
Besides literals~$\ell_j$,
a $\# minimize$ statement includes integer weights $w_j$ for $1\leq j\leq k$.
A $\# minimize$ statement distinguishes optimal answer sets of a program
as the ones yielding the smallest weighted sum for the true literals among
$\ell_1,\dots,\ell_k$.
For a formal introduction,
we refer the interested reader to~\cite{siniso02a}.

Likewise, first-order representations, commonly used to encode problems in ASP,
are only informally introduced.
In fact,
\gringo\ requires programs to be \emph{safe}, that is, each variable must occur
in a positive body literal.
Formally, 
we only rely on the function $\mathit{ground}$ to denote the set of all ground
instances, $\mathit{ground}(\Pi)$, of a program $\Pi$ containing first-order variables.
Further language constructs of interest,
include conditional literals, like ``$a\!:\!b$'',
the range and pooling operator ``..'' and ``;''
as well as standard arithmetic operations.
The ``:'' connective expands to the list of all instances of its left-hand side
such that corresponding instances of literals on the right-hand side
hold~\cite{lparseManual,potasscoManual}.
While ``..'' allows for specifying integer intervals,
``;'' allows for pooling alternative terms to be used as arguments within an
atom.
For instance, $p(1..3)$ as well as $p(1;2;3)$ stand for the three facts $p(1)$,
$p(2)$, and $p(3)$.
Given this, $q(X)\!:\!p(X)$ results in $q(1), q(2), q(3)$.
See \cite{potasscoManual} for a detailed description of the input language of the
grounder \gringo.

%%% Local Variables: 
%%% mode: latex
%%% TeX-master: "paper"
%%% End: 

\setcounter{equation}{0}
\section{Declarative Automatic Network Reconstruction}\label{sec:approach}

Our approach expresses the ANR Problem in form of a logic program
under answer set semantics.
In the next sections, we explain the encoding in detail,
starting with the representation of the experiment graph.
\subsection{Representing experiment instances}
As common in ASP, a problem instance is given as facts.
We define the instance of an experiment graph $(X,\PerturbationS\cup\ResponseS)$ over \SpecieS\ with domain $D$,
as a set of facts
\[
\begin{array}{rcl}
\Instance(X,\PerturbationS\cup\ResponseS) & = &\{\species{s}\mid s \in S\}
\\
&\cup&\{\capacity{s_i}{d_i}\mid s_i\in S\ \text{and}\ d_i \in D\}
\\
&\cup&\{\state{x}\mid x\in X\}
\\
&\cup&\{\perturbation{x}{x'}\mid (x,x')\in \PerturbationS\}
\\
&\cup&\{\response{x}{x'}\mid (x,x')\in \ResponseS\}
\\
&\cup&\{\terminalstate{x'}\mid x'=t(x), x\in X\}
\\
&\cup&\{\amount{x}{s_i}{x_i} \mid x=(x_1,\dots,x_n)\in X, 1 \leq i \leq n\}.
\end{array}
\]
Predicates $\species{s_i}$ and $\capacity{s_i}{d_i}$ denote the species $s_i \in \SpecieS$ over their associated
capacities $d_i\in D$.
Similarly, we use $\state{x}$ to denote each state $x\in X$ and mark terminal states
with predicate $\terminalstate{x}$.
For the edges of the graph, we use predicate $\edge{T}{x}{x'}$,
where $T$ can be either $p$ or $r$ to indicate that $(x,x')$ is a perturbation or
a response edge, respectively.

As an example,
Table~\ref{tab:exp:graph} gives the specification of the experiment graph in
Figure~\ref{figexpgraph}.
% ------------------------------------------------------------
\begin{table}[ht]
  \centering
\progtest{lll}{
\fact{\species{fr}}
&
\fact{\species{r}}
&
\fact{\species{spo}}
\\
\fact{\capacity{fr}{0..1}}
\qquad
&
\fact{\capacity{r}{0..1}}
\qquad
&
\fact{\capacity{spo}{0..1}}
\nonumber
}

\progtest{llll}{
\fact{\state{x^0}}
&
\fact{\state{x^1}}
&
\fact{\state{x^2}}
&
\fact{\state{x^3}}
}
\vspace{-12pt}
\progtest{lllll}{
\fact{\state{x^4}}
&
\fact{\state{x^5}}
&
\fact{\state{x^6}}
&
\fact{\state{x^7}}
&
\fact{\state{x^8}}
}

\progtest{lll}{
\fact{\perturbation{x^0}{x^1}}
\qquad
&
\fact{\perturbation{x^2}{x^5}}
\quad
&
\fact{\perturbation{x^3}{x^6}}
\\
\fact{\response{x^1}{x^2}}
&
\fact{\response{x^2}{x^3}}
&
\fact{\response{x^3}{x^4}}
\\
\fact{\response{x^5}{x^0}}
&
\fact{\response{x^6}{x^7}}
&
\fact{\response{x^7}{x^8}}
}

\progtest{lll}{
\fact{\terminalstate{x^0}}
\qquad
&
\fact{\terminalstate{x^4}}
\qquad
&
\fact{\terminalstate{x^8}}
}

\progtest{llll}{
\fact{\amount{x^0}{fr}{0}}
\quad
&
\fact{\amount{x^1}{fr}{1}}
\quad
&
\fact{\amount{x^2}{fr}{0}}
&
\fact{\amount{x^3}{fr}{0}}
\\
\fact{\amount{x^0}{r}{0}}
&
\fact{\amount{x^1}{r}{0}}
&
\fact{\amount{x^2}{r}{0}}
&
\fact{\amount{x^3}{r}{0}}
\\
\fact{\amount{x^0}{spo}{0}}
&
\fact{\amount{x^1}{spo}{0}}
&
\fact{\amount{x^2}{spo}{0}}
&
\fact{\amount{x^3}{spo}{0}}
}
\progtest{@{}l@{}l@{}l@{}l@{}l@{}}{
\fact{\amount{x^4}{fr}{0}}
&
\fact{\amount{x^5}{fr}{0}}
&
\fact{\amount{x^6}{fr}{0}}
&
\fact{\amount{x^7}{fr}{0}}
&
\fact{\amount{x^8}{fr}{0}}
\\
\fact{\amount{x^4}{r}{0}}
&
\fact{\amount{x^5}{r}{1}}
&
\fact{\amount{x^6}{r}{1}}
&
\fact{\amount{x^7}{r}{0}}
&
\fact{\amount{x^8}{r}{0}}
\\
\fact{\amount{x^4}{spo}{1}}
&
\fact{\amount{x^5}{spo}{0}}
&
\fact{\amount{x^6}{spo}{0}}
&
\fact{\amount{x^7}{spo}{0}}
&
\fact{\amount{x^8}{spo}{1}}
}
  \caption{Specification of Experiment Graph in Figure~\ref{figexpgraph}.}
  \label{tab:exp:graph}
\end{table}

%%% Local Variables: 
%%% mode: latex
%%% TeX-master: "../paper"
%%% End: 

% ------------------------------------------------------------
Given the instance in Table~\ref{prog:det}, the solutions of our final logic program 
represent all regulatory structures that are conformal with the extensions of the experiment graph.
We start with checking the validity of the experiment graph.

\subsection{Checking Validity}

In Section \ref{sec:problem}, three conditions were specified for validity.
Checking these conditions can be done with the following logic program.

Condition \textbf{I},
ensuring that each state has only one outgoing arc,
is given in~\eqref{prog:det}.
\prog{
\impl{}{\response{X_1}{X_2}, \response{X_1}{X_3}, X_2 \neq X_3}
\label{prog:det}
}
Rules \eqref{prog:determinismstart} to \eqref{prog:determinismend} account
for Condition \textbf{II}.
We first collect all pairs of states that are not equal
and then compute the associated terminal state of each state which can be determined deterministically.
Rule~\eqref{prog:determinismend}
ensures that no two equal states lead to unequal terminal states.
% ------------------------------------------------------------
\prog{
\impl{\unequal{X_1}{X_2}}{\amount{X_1}{S}{V}, \naf{\amount{X_2}{S}{V}}, \state{X_2}}
\label{prog:determinismstart}
\\
\impl{\associatedTerminal{X}{T}}{\response{X}{T}, \terminalstate{T}}
\\
\impl{\associatedTerminal{X_1}{T}}{\response{X_1}{X_2}, \associatedTerminal{X_2}{T}}
\label{prog:associatedterminalend}
\\
\impl{}{\naf{\unequal{X_1}{X_2}}, \unequal{T_1}{T_2}, \associatedTerminal{X_1}{T_1}, \associatedTerminal{X_2}{T_2}}
\label{prog:determinismend}
}
% ------------------------------------------------------------
The rules in \eqref{prog:decreasestart} and \eqref{prog:decreaseend} ensure a decrease in each response
as required in Condition~\textbf{III}.
\prog{
\impl{\decreaseone{X_1}}{\response{X_1}{X_2}, \amount{X_1}{S}{V_1}, \amount{X_2}{S}{V_2},
\label{prog:decreasestart}
\\
&V_2-V_1 < 0}
\\
\impl{}{\naf{\decreaseone{X_1}}, \response{X_1}{X_2}}
\label{prog:decreaseend}
}
% ------------------------------------------------------------
The next proposition ensures correctness and completeness of this logic program.
\begin{proposition}
Let $(X,\PerturbationS\cup\ResponseS)$ be the experiment graph
and 
$\Pi_1$ be the logic program $ground(\Instance(X,\PerturbationS\cup\ResponseS)\cup\{\eqref{prog:det},\dots,\eqref{prog:determinismend}\})$.

Then,
the experiment graph $(X,\PerturbationS\cup\ResponseS)$ is valid
iff there exists an answer set of $\Pi_1$.
\end{proposition}
The proof of this and all following results follow from the construction of the
respective logic programs.
Recall that the experiment graph in Figure~\ref{figexpgraph} is invalid,
so the corresponding program has no answer set.

\subsection{Building Regulatory Structures}

We now proceed by defining a finite logic program that allows us to find all
regulatory structures conformal with a valid experiment graph.

For guaranteeing the finiteness of the ground program,
we need to know the maximum number of reactions that shall be used.
As each reaction has to consume at least one species, the number of reactions
is bound by the total number of decreases during a response edge.
Although there exist better approximations, for simplicity,
we define the logic program $\Bound(X,\PerturbationS\cup\ResponseS)$ that
results in the single answer set $A$,
such that $\maxadds{n}\in A$, $\maxreactions{m}\in A$ and
$\responsethree{x}{x'}{m}\in A$ for all $(x,x')\in\ResponseS$,
where $n,m\in\mathds{N}_0$ are sufficient bounds.
%we expect that the maximum number of reactions is given via the predicate
%\emph{\maxreactionsn}.
% ------------------------------------------------------------
%\prog{
%\impl{\numreactions{M}}{M = sum [\change{X_1}{X_2}{S}{V} = -V : V < 0]}
%\label{prog:numreactionsone}
%\\
%\impl{\maxreactions{M}}{M = sum [\numreactions{N}]}
%\label{prog:numreactionstwo}
%}
% ------------------------------------------------------------
Given these bounds, we now choose a certain number of reactions in \eqref{prog:chosereaction}
to be part of the regulatory structure.
% ------------------------------------------------------------
\prog{
%\fact{\maxreactions{maxReaction}}
%\\
\fact{\reactionone{1}}
\\
\impl{\{\;\reactionone{N{+}1}\;\}}{\reactionone{N}, \maxreactions{M},N {<} M}
\label{prog:chosereaction}
}

For the resulting reaction vector, we pick out its values in \eqref{prog:reactionvalues}.
So, for each reaction and each species $s_i\in\SpecieS$,
we choose exactly one value from the set $\{d,-d \mid d\in D_i\}$. %  for the reaction vector.
This is constrained by \eqref{prog:consume} ensuring that each reaction has at least one negative entry.
\prog{
%\impl{\capacity{S}{X-1}}{\capacity{S}{X}, X > 0}
%\\
%\impl{\range{S}{V;-V}}{\capacity{S}{V}}
%\\
%\impl{1\ \{\;\reactionthree{R}{S}{V;{-}V} : \capacity{S}{V}\;\}\ 1}{\reactionone{R}, \species{S}}
%\\
\impl{1\ \{\;\reactionthree{R}{S}{V} : \capacity{S}{V},\hspace{11.4mm}\\
\reactionthree{R}{S}{{-}V} : \capacity{S}{V}\}\ 1}{\reactionone{R}, \species{S}}
\label{prog:reactionvalues}
\\
%does not consume
\impl{}{\{\;\reactionthree{R}{S}{V} : V < 0\;\}\ 0,\reactionone{R}}
\label{prog:consume}
}

Next, we want to define the realizing sequences guessing a partial order \PO\ of reactions.
Therefore, we define the intermediate states of a sequence.
Each consecutive intermediate state is built, adding the currently fastest enabled reaction.
%The fastest reaction is freely chosen between the applicable ones.
%We again assume that we have given the maximum number of reactions in each sequence, via the predicate
%\emph{\responsethree{X_1}{X_2}{L}}.
% ------------------------------------------------------------
%\prog{
%\impl{\hspace*{-4mm}\numreactionsthree{X_1}{X_2}{M}}{M {=} sum [\change{X_1}{X_2}{S}{V} {=} {-}V : V {<} 0],
%\\
%&\response{X_1}{X_2}}
%\label{prog:reactionspersequenceone}
%\\
%\impl{\maxreactionsthree{X_1}{X_2}{M}}{M = sum [\numreactionsthree{X_1}{X_2}{N}],
%\\
%&\response{X_1}{X_2}}
%\label{prog:reactionspersequencetwo}
%}
% ------------------------------------------------------------
As with the reactions, we first guess the length of the sequences $\sigma(x,x')$
in \eqref{prog:lengthstart} and \eqref{prog:lengthend},
which is bound by the precomputed predicate \responsethree{x}{x'}{.}.
\prog{
\impl{\step{X_1}{0}}{\response{X_1}{X_2}}
\label{prog:lengthstart}
\\
\impl{\{\;\step{X_1}{N+1}\;\}}{\step{X_1}{N}, \responsethree{X_1}{X_2}{M}, N < M}
\label{prog:lengthend}
}
Then, the value of the species is defined for each intermediate step
by Rule \eqref{prog:intermediatestart} and \eqref{prog:intermediateend}
by adding the fastest reaction, as stated in Condition \textbf{IV}.
\prog{
\impl{\intervalue{0}{X}{S}{V}}{\amount{X}{S}{V}}
\label{prog:intermediatestart}
\\
\impl{\intervalue{L+1}{X}{S}{V_{Old}+V_{New}}}{\intervalue{L}{X}{S}{V_{Old}},
\label{prog:intermediateend}
\\
&\fastest{R}{L}{X}, \reactionthree{R}{S}{V_{New}},
\\
&\capacity{S}{V_{Old}+V_{New}},
\\
&\step{X}{L}}
}
Rules \eqref{prog:applicablestart} and \eqref{prog:applicableend}
find out which reaction is enabled in which intermediate state.
\prog{
%%%%%%%%%%%%%%%%%%%%%%%%%%%%%%%%%%%%%%%%%%%%%%%%%%%%%%%%%%%%%%%%
% If a reaction is applicable in an intermediate step, but
% does not apply, it must be slower the reaction that applies
%%%%%%%%%%%%%%%%%%%%%%%%%%%%%%%%%%%%%%%%%%%%%%%%%%%%%%%%%%%%%%%%
\impl{\napplicable{R}{L}{X}}{\intervalue{L}{X}{S}{V_1}, \reactionthree{R}{S}{V_2}, \species{S},
\label{prog:applicablestart}
\\
&\naf{\capacity{S}{V_1+V_2}}, \step{X}{L}}
\\
\impl{\applicable{R}{L}{X}}{\naf{\napplicable{R}{L}{X}},\reactionone{R}, \step{X}{L}}
\label{prog:applicableend}
}
From the enabled reactions, we freely choose a fastest reaction via Rule \eqref{prog:fastest}.
\prog{
\impl{\{\;\fastest{R}{L}{X}\;\}}{\applicable{R}{L}{X}}
\label{prog:fastest}
}
To impose an ordering on the reactions, Rule~\eqref{prog:nfaster} says that
if there is another enabled reaction different from the fastest one, 
then this one must be slower.
\prog{
\impl{\slower{R_2}{R_1}} {\fastest{R_1}{L}{X_1},
\applicable{R_2}{L}{X_1}, R_1 \neq R_2}
\label{prog:nfaster}
}
We just need to add transitivity to the predicate \emph{\slowern},
and forbid that a reaction is slower than itself % (Line \eqref{prog:slowerstart} and \eqref{prog:slowerend})
to enforce the partial order:
\prog{
\impl{\slower{R_1}{R_3}}{\slower{R_1}{R_2}, \slower{R_2}{R_3}}
\label{prog:slowerstart}
\\
\impl{}{\slower{R}{R}, \reactionone{R}}
\label{prog:slowerend}
}
To create a valid realizing sequence, 
the constraint in~\eqref{prog:nextstate} assures that the sequence of reactions 
leads to the next measured state of the system, as stipulated in~\textbf{V}.
\prog{
%reach next state
\impl{}{\intervalue{M+1}{X_1}{S}{V}, \response{X_1}{X_2},\naf{\amount{X_2}{S}{V}}, 
\label{prog:nextstate}
\\
&\step{X}{M}, \naf{\step{X}{M+1}},\species{S}}
}
Similarly, \eqref{prog:monotonestart} and \eqref{prog:monotoneend} enforce
that reactions are monotone, as dictated by Condition~\textbf{VI}.
This is expressed by saying that a reaction applying in a state $x$ must
increase/decrease the species into the direction of the next state $x'$, 
where $(x,x')\in\ResponseS$.
\prog{
%monotonocity
\impl{}{\fastest{R}{L}{X_1}, \reactionthree{R}{S}{V_3}, \response{X_1}{X_2}, 
\label{prog:monotonestart}
\\
&\amount{X_1}{S}{V_1}, \amount{X_2}{S}{V_2}, (V_2-V_1)*V_3<0}
\\
\impl{}{\fastest{R}{L}{X_1}, \naf{\reactionthree{R}{S}{0}}, \response{X_1}{X_2},
\label{prog:monotoneend}
\\
&\amount{X_1}{S}{V}, \amount{X_2}{S}{V}}
}
Condition \textbf{VII} states that for creating a \emph{regulatory structure},
no reaction my be enabled in a terminal state.
This is addressed in \eqref{prog:appterminalstart} and \eqref{prog:appterminalend}:
\prog{
%inapplicable in terminal state
\impl{\napplicableterminal{R}{X}}{\terminalstate{X}, \reactionthree{R}{S}{V_1},
\label{prog:appterminalstart}
\\
&\amount{X}{S}{V_2}, \naf{\capacity{S}{V_1+V_2}}}
\\
% }
% \prog{
\label{prog:appterminalend}
\impl{}{1\ \{\;\naf{\napplicableterminal{R}{X} : \terminalstate{X}}\;\}, \reactionone{R}}
}

To avoid irrelevant solutions, every reaction must be used.
This is addressed in conditions \textbf{VIII} and \textbf{IX}.
That is, each reaction has to be at least one time the
fastest reaction (see \eqref{prog:oncefasteststart})
and in each step there must exist at least one fastest
reaction (see \eqref{prog:oncefastestend}).
\prog{
%every reaction must be at least one time the fastest
\impl{}{\reactionone{R}, \naf{\ 1\ \{\;\fastest{R}{L}{X}\;\}}}
\label{prog:oncefasteststart}
\\
%every step needs a fastest reaction
\impl{}{\step{X}{L}, \naf{\ 1\ \{\;\fastest{R}{L}{X}\;\}}}
\label{prog:oncefastestend}
}
% ------------------------------------------------------------

For formulating our correctness and completeness result, we need the following auxiliary definition.
\begin{definition}
Let $(s_1,\dots,s_n)$ be the vector of the species in \SpecieS.
Let $A$ be a set of ground atoms such that $\reactionone{x}\in A$ and 
$\reactionthree{x}{s_i}{v_i}\in A$ for $1 \leq i \leq n$.
Then, define $\ReactionVector{A}{x}=(v_1,\dots,v_n)$.
\end{definition}
We then get the following result.
\begin{proposition}
Let $(X,\PerturbationS\cup\ResponseS)$ be the experiment graph and $\Pi_2$ be the logic program
$ground(\Instance(X,\PerturbationS\cup\ResponseS)\cup\Bound(X,\PerturbationS\cup\ResponseS)\cup\{\eqref{prog:det},\dots,\eqref{prog:oncefastestend}\})$.
\begin{itemize}
\item If $A$ is an answer set of $\Pi_2$,
then the regulatory structure $(\Network,\PO)$ is conformal with the experiment graph $(X,\PerturbationS\cup\ResponseS)$,
where $\Network=\{\ReactionVector{A}{x}\mid\reactionone{x}\in A\}$ and
$\PO=\{(\ReactionVector{A}{x},\ReactionVector{A}{x'})\mid \slower{x}{x'}\in A\}$.
\item If there exists a regulatory structure $(\Network,\PO)$
that is conformal with the experiment graph $(X,\PerturbationS\cup\ResponseS)$,
then there exists an answer set $A$ of logic program $\Pi_2$
such that $\Network=\{\ReactionVector{A}{x}\mid\reactionone{x}\in A\}$ and
$\PO=\{(\ReactionVector{A}{x},\ReactionVector{A}{x'})\mid \slower{x}{x'}\in A\}$.
\end{itemize}
\end{proposition}

\subsection{Recovering the Experiment Graph}
We now extend our logic program to recover from an invalid experiment graph.

Given that the minimum number of additional species is provided via predicate \emph{\maxadds{M}},
we start by introducing $M$ additional species and their capacities: % (Line \eqref{prog:addspeciesstart} and \eqref{prog:addspeciesend}).
% ------------------------------------------------------------
\prog{
\impl{\addspecies{1..M}}{\maxadds{M}}
\label{prog:addspeciesstart}
\\
\impl{\capacity{S}{V}}{\addspecies{S}, V=0..1}
\label{prog:addspeciesend}
}
We freely choose a value for the additional species in \eqref{prog:addvalue}.
Moreover,
we declare them as ordinary species in \eqref{prog:addtonormal} in order
to subject them to all constraints on species.
\prog{
\impl{1\ \{\;\amount{X}{S}{V} : \capacity{S}{V}\;\}\ 1}{\addspecies{S}, \state{X}}
\label{prog:addvalue}
\\
\impl{\species{S}}{\addspecies{S}}
\label{prog:addtonormal}
}
In a valid extension, 
there may be no change in the additional species during a perturbation,
according to Condition~\textbf{XI}:
\prog{
\impl{}{\perturbation{X_1}{X_2}, \amount{X_1}{S}{V}, \naf{\amount{X_2}{S}{V}}, \addspecies{S}}
\label{prog:addnochange}
}
Finally,
for minimizing the number of changes in the additional species,
captured in \eqref{prog:minimizestart}, 
we use the minimize statement in \eqref{prog:minimizeend}.
\prog{
\impl{\addchangeone{X_1}}{\response{X_1}{X_2}, \amount{X_1}{S}{V},
\label{prog:minimizestart}
\\
&\naf{\amount{X_2}{S}{V}}, \addspecies{S}}
\\
\fact{\# minimize \{\;\addchangeone{X}\;\}}
\label{prog:minimizeend}
}

As before, we get the following correctness and completeness result for this
encoding.
%
% ------------------------------------------------------------
\begin{proposition}
Let $(X,\PerturbationS\cup\ResponseS)$ be the experiment graph and $\Pi_3$ the logic program 
$ground(\Instance(X,\PerturbationS\cup\ResponseS)\cup\Bound(X,\PerturbationS\cup\ResponseS)\cup\{\eqref{prog:det},\dots,\eqref{prog:minimizestart}\})$.
\begin{itemize}
\item If $A$ is an answer set of $\Pi_3$
being minimal wrt statement $ground(\eqref{prog:minimizeend})$,
the regulatory structure $(\Network,\PO)$ is conformal with a minimal valid extension%
\footnote{Recall that such an extension is minimal wrt to the number of species and changes in the additional species.}
of the experiment graph $(X,\PerturbationS\cup\ResponseS)$,
where
$\Network=\{\ReactionVector{A}{x}\mid\reactionone{x}\in A\}$ and
$\PO=\{(\ReactionVector{A}{x},\ReactionVector{A}{x'})\mid \slower{x}{x'}\in A\}$.
\item If $(\Network,\PO)$ is a regulatory structure that is conformal with a minimal valid extension
of the experiment graph $(X,\PerturbationS\cup\ResponseS)$, 
then there is an answer set $A$ of $\Pi_3$ 
being minimal wrt statement $ground(\eqref{prog:minimizeend})$
such that
$\Network=\{\ReactionVector{A}{x}\mid\reactionone{x}\in A\}$ and
$\PO=\{(\ReactionVector{A}{x},\ReactionVector{A}{x'})\mid \slower{x}{x'}\in A\}$.
\end{itemize}
\end{proposition}
Given the ground instance $I$ in Table~\ref{prog:det},
which is the logic representation of the experiment graph in Figure~\ref{figexpgraph}.
The answer sets of 
\begin{align*}
\mathit{ground}(& I \cup \{\eqref{prog:det},\dots,\eqref{prog:minimizestart}\} \cup \{\maxreactions{12}\}\cup\{\maxadds{2}\}\ \cup
\\
&
\{\responsethree{x_1}{x_2}{1}, \responsethree{x_2}{x_3}{1}, \responsethree{x_3}{x_4}{1},
\\
&\phantom{\{} \responsethree{x_7}{x_8}{2}, \responsethree{x_6}{x_7}{2}, \responsethree{x_5}{x_0}{3} \})
\end{align*} 
minimal wrt to 
the statement $ground(\eqref{prog:minimizeend})$ do correspond to the
regulatory structures shown in Figure~\ref{fig:sol:one} and \ref{fig:sol:two}.
Actually, there are four answer sets with symmetric behavior on the additional species.
To avoid these, we use a symmetry breaking technique, that is explained in the next section.

%$ground(I\cup\{\eqref{prog:det},\dots,\eqref{prog:minimizestart}\}\cup\{\maxreactions{12}.\}\cup\{\maxadds{2}.\}
%\cup\{\fact{\responsethree{x_1}{x_2}{1}}, \fact{\responsethree{x_2}{x_3}{1}}, \fact{\responsethree{x_3}{x_4}{1}},
%\fact{\responsethree{x_7}{x_8}{2}}, \fact{\responsethree{x_6}{x_7}{2}}, \fact{\responsethree{x_5}{x_0}{3}} \})$ 
\subsection{Symmetry breaking}
To avoid symmetric models and to speed up computation, 
we developed symmetry breaking rules for the additional species and the reactions.
The additional species can be freely labeled obeying the constraints.
Therefore producing $2^a$ extensions of the experiment graph.
Given the conditions that must hold for a valid experiment graph and a conformal regulatory structure,
the number of extensions is restricted.
Despite all that, unnecessary extensions can be created, as for each extension in
Figure \ref{fig:ext:one} and \ref{fig:ext:two} a mirrored version exists
where the labeling of the additional species is switched between additional species $x$ and $y$.
To overcome this, we define an order on the states.\footnote{Any total order is valid.}
The evolution of the added species $s$ is considered to ``precede'' that of added species $s+1$,
if either $s$ changes and $s+1$ not, or if $s$ decreases more than $s+1$
(increases handled analogously).
The omitted models can easily be reconstructed by permuting the added species.
This is of course not necessary, as a biologist has to research the ``meaning'' of the species.

Furthermore, we do symmetry breaking on reactions.
As we name reactions by numbers, we enforce them to respect some order.\footnotemark[\value{footnote}]
We simply use the reaction vector of each reaction to impose a fixed order.
So the reaction with the identifier $1$ always has the ``smallest'' reaction vector.
This way we omit models that differ only in the naming of the reactions.

As with the refinements of our encoding,
discussed in the next section,
the corresponding logic programs can be found at~\cite{website}.

\subsection{Refinements}
\label{sec:refinements}
The encoding that we have presented above was optimized for readability.
An enhanced version optimized for performance can be found on the web~\cite{website}.
Also, it contains an optimized possibility to compute the static bounds:
the maximum number of reactions, additional species and number of possible intermediate states.
A basic approximation for the number of reactions is the number of negative changes of each species,
as each reaction has to consume at least one species.
This usually results in a high maximum number of reactions.
As this number is crucial for the systems performance,
we first solve the problem without checking the partial order of the reactions
and maximize the number of reactions that shall be used.
This computation can be done much faster and gives good approximations for the
maximum number of reactions.
For approximating the number of intermediate steps,
we have a similar approach, 
considering only two consecutive states and again maximizing the number
of reactions that are enabled in between.

We now describe another interesting optimization,
this time for the encoding itself.
To reduce the size of the strongly connected components of the 
positive dependency graph~\cite{linzha04a} of the logic program,
we are using a complete ordering of the reactions instead of a partial one.
To this end, we replace \eqref{prog:nfaster} with the following rules:
\prog{
\impl{\{\;\slower{R_1}{R_2}\;\}}{\reactionone{R_1},\reactionone{R_2}, R_1 \neq R_2}
\label{prog:ref1}
\\
\impl{}{\fastest{R_1}{L}{X}, \applicable{R_2}{L}{X},
\\
&R_1 \neq R_2, \naf{\slower{R_2}{R_1}}}
\label{prog:ref2}
}
We now freely choose an ordering of the reactions in \eqref{prog:ref1}
and then assure in \eqref{prog:ref2} that each other enabled reaction has been chosen to be slower.
In this way, we reduce the size of the positive cycles in the dependency graph.
Now we are no longer restricted to partial orders and the number of different solutions would
increase drastically, as each partial order implies many orderings.
To overcome this issue, we project only on the reaction vectors.
This means that we compute all (minimal) solutions that differ in the reaction vectors
but avoid solutions with the same set of reaction vectors but different orderings.
This is a feature of our solver \clasp{} and can be done very efficiently
as shown in~\cite{gekasc09a} without enumerating all solutions.
Furthermore, different redundant constraints have been added to the encoding.
For example, a response may not be enabled in a terminal state,
reactions that apply in a state must sum up to the difference vector, etc.
Without these optimizations of the encoding,
for instance,
we were unable to solve the instance ``ip3r-1-4-dag'' used in the next section.
%As we are only interested in some features of the models, we use the \emph{projection}
%feature of our ASP solver \clasp.
%It allows us to project on a set of predicates, avoiding the computation
%of all models that do not differ on this set.
%On enumerating all solutions, we project on the set of reaction vectors,
%omitting all models that do have the same network and only differ in the ordering of reactions,
%as there might exist many orderings.
%\comment{M: This is not the case with the described encoding}

%%% Local Variables: 
%%% mode: latex
%%% TeX-master: "paper"
%%% End: 

\section{Experiments}\label{sec:experiments}
To test the feasibility of our approach,
we used ``in-silico''\footnote{The data is confidential and was made anonymous by
  our industrial partner.} 
experiments generated from a synthetic bio-chemical network.
Several time series are generated and some combinations of them are shown in Table~\ref{tab:benchmarks}.
From the time series, the values of two species were removed to simulate experiments not measuring all species.
So the data of 14 species is used.
We compared our ASP approach to the direct implementation described in~\cite{preprint}.
Unfortunately,
no direct implementation exists that does handle the range of constraints described here.
Some approaches additionally handle the creation of catalysators/inhibitors and others lack the check of 
the partial ordering on the reactions.
The implementation that comes closest to ours does not compute the number of additional species
(it has to be given from outside) and does not do the partial order check.
But for the feasibility test, we decided to compare with this version,
referred to as the ``direct implementation''.
We tested it, giving the maximum number of additional species, as computed by our approach, as input.
The benchmark were run single-threaded on 
an Intel Xeon machine with 32GB main memory  possessing two 3,4Ghz
processor with eight cores each;
each benchmark was restricted to 2GB of memory and 1h runtime.
For the times, we show the average of three runs in seconds.
MEM indicates that the memory limit was reached.
For the ASP approach we use the grounder \gringo~(3.0.4) and the solver \clasp~(1.3.6).
We tested the unoptimzed version (denoted by ``unopt. asp times'') of our encoding as well as the 
optimzed version with the refinements from Section~\ref{sec:refinements} (denoted by ``asp times'').
\begin{table}
\begin{tabular}{ |l | r | r | c| c| c| c|c|c|}
\cline{1-9}
 & ASP & unopt. & direct impl. & add. &  & maximum & &exper-\\
instances & times & \hspace{-2mm}ASP times& times& \hspace{-2mm}species & \hspace{-2mm}models & \hspace{-2mm}of reactions & \hspace{-2mm}states & \hspace{-2mm}iments\\
\cline{1-9}
	ip3r-1& 1.3 & 4.5 & 0.1 &1 & 4 & 7 & 11 &  2\\
\cline{1-9}
	ip3r-1-dag& 1.1 & 4.2 & 0.1 &1 & 4 & 6 & 12 &  3\\
\cline{1-9}
	ip3r-2-dag& 0.7 & 1.5 & 0.1 &1 & 2 & 4 & 11 &  4\\
\cline{1-9}
	ip3r-3-dag& 0.3 & 0.5 & 0.1 &0 & 1 & 4 & 11 &  4\\
\cline{1-9}
	ip3r-4-dag& 0.5 & 1.0 & 0.1 &1 & 8 & 5 & 10 &  4\\
\cline{1-9}
	ip3r-1+4-dag& 3.6 & 16.4 & 0.1 &1 & 8 & 8 & 20 &  6\\
\cline{1-9}
	ip3r-1+2-dag& 34.0 & 300.9 & MEM &2 & 44 & 10 & 21 &  6\\
\cline{1-9}
	ip3r-1+3-dag& 59.0 & 750.7 & MEM&2 & 128 & 10 & 21 &  6\\
\cline{1-9}
  ip3r& 30.2 & 244.6 & MEM &1 & 2 & 9 & 37 &  11\\
\cline{1-9}
	ip3r-1-4& 656.7 & 3435.3 & MEM &2 & 104 & 11 & 37 &  11\\
\cline{1-9}
	ip3r-1-4-dag&\hspace{-3mm}3562.3 & TIME & MEM &2 & 280 & 12 & 38 &  12\\
\cline{1-9}
\end{tabular}
\caption{Reconstructing a model using in-silico experiments}
\label{tab:benchmarks}
\end{table}
The number of models gives the number of different regulatory structures that have been reconstructed by the ASP approach.
The number of states (measured time points) and experiments used for the reconstruction is also given.
With ``maximum of reactions'',
we refer to the maximum of reactions that are used in the regulatory structures.
As it can easily be seen in Table~\ref{tab:benchmarks}, the number of experiments 
(and therefore the number of states)
and additional species
increases the difficulty of the problem.
Our refined approach was able to solve all of the problems,
which means that it is feasible to run it on the shown number of experiments and states,
as long as the number of additional species stays low.
The direct implementation, also having limited functionality, has severe problems with memory usage.
\footnote{We also tested it with 3GB memory restriction, which did not changed any of the results.}
On the other hand, it has less initialization overhead on the small examples.

%%% Local Variables: 
%%% mode: latex
%%% TeX-master: "paper"
%%% End: 

\section{Discussion}\label{sec:discussion}
In the area of automatic network reconstruction many different approaches have been developed.
They differ in the used techniques and the kind of system they reconstruct.
Statistical methods are used e.g. in \cite{gifjaa01a} and \cite{pereelfr01a}
reconstructing wiring diagrams using Bayesian network methods.
More descriptive systems are time continuous deterministic dynamical systems.
Using ordinary differential equations, \cite{yeteco02a} and \cite{lausti04a}
infer a network by solving a non-homogeneous system of linear equations, given a set of experiments.
The result is a minimal network in means of the structure of the functions.
Enumerating algorithms are used in \cite{lifuso98a} and \cite{akmiku99a}
to search for the sparsest Boolean model.
Boolean networks however are more intuitive but less expressive.
Hybrid models seem to compensate the drawbacks of Boolean models.
\cite{relian02a} uses a genetic algorithm on gene expression data to produce a hybrid model,
including quantitative and qualitative information.
Depending on the quality of the available experimental data and the 
type of the studied models, further approaches have been developed;
see~\cite{preprint,duwawema08a,wagler11a} for a detailed comparison to our underlying approach 
in Section~\ref{sec:problem}.

Our approach does not try to find the ``best'' model, because this approach looses information
about important alternatives that only a biologist can decide on.
It rather shows \emph{all} possible models conform with the experiments.
We use a hybrid model, incorporating several discrete levels of concentrations for the species.
This extends the purely Boolean approach, but can of course not
keep up with the expressiveness of differential equations.
The modeling as a logic program makes it simple to integrate all the various constraints.
The approach can easily be extended by further constraints, or constraints can be relaxed.
Although we used a state-of-the-art solver for logic programs, we rigorously had to shrink
the size of the problem due to different preprocessing steps.
In contrast to other approaches, this does not change the problem or the solutions.
We still infer all possible explanations for the experiments.
Furthermore our approach benefits from future developments in ASP solving,
like parallelization.

From a broader perspective,
ASP has already proved its utility for diverse biological applications.
Among them,
we find
\cite{bachtrtrjobe04a,dwgrniscse07a,gescthusve08b,erdtur08a,schthi09a,erdem09a,dadopo09a,geguivscsithve10a},
all of which treat rather different biological problems from what we tackled in
the paper at hand.
A feature common to many among these approaches is the exploitation of ASP's
combinatorial nature in inspecting either all or what is common to all solutions
to a biological problem.

%Various approaches:
%\begin{itemize}
%\item statistical methods DONE
%\item reconstructing only the wiring diagram using Bayesian network methods DONE
%   \begin{itemize}
%	 \item \cite{gifjaa01a} DONE
%	 \item \cite{pereelfr01a} DONE
%	 \end{itemize}
%\item  other
%\begin{itemize}
%\item using expression profiles
%\item \cite{gabeloco03a}
%\end{itemize}
%time continious deterministic dynamical systems, using ordinary differential equations
%\begin{itemize}
%\item \cite{yeteco02a}
%\item finds the sparset consistent model
%\item \cite{lausti04a}
%  \begin{itemize}
%   \item time discrete multi-state dynamical systems
%   \item given a set of experiments, if infers a network by solving the solutions to a non-homogeneous system of linear equations
%   \item a minimal network in means of the structure of the functions
%  \end{itemize}
%Boolean networks
%\begin{itemize}
%\item more intuitive, less expressive
%\item REVEAL \cite{lifuso98a}, optimal network done with fewest variables,
%\item \cite{akmiku99a}, simple model
%\item both approaches restrict the inputs of the reactions
%
%\end{itemize}
%Hybrid models
%\begin{itemize}
%\item \cite{relian02a} using genetic algorithm on gene expression data
%\end{itemize}
%\cite{brasch03a} proposing a new hybrid model
%\end{itemize}
%\end{itemize}

%%% Local Variables: 
%%% mode: latex
%%% TeX-master: "paper"
%%% End: 

\section{Summary}\label{sec:summary}
We presented a declarative solution to the ANR problem using ASP.
We support checking validity of an experiment graph,
predicting the behavior of unmeasured species,
and 
reconstructing all possible explanations for a given set of experiments
using a partial order on the used reactions.
We showed that the mathematical representation of the problem can be easily
translated into a logic program which then can be handled by a state of the art
grounder and solver for ASP.
As the logic program can easily be split into different parts,
also various versions of the problem (adding or relaxing some of the constraints)
can be tackled.
This is especially useful when the problem is refined to use catalysts or inhibitors
as done in~\cite{catalysts}
or just has to be changed for special purposes.
It can also be used to introduce P-Invariants as described in~\cite{preprint}.
This avoids rewriting complex programming code and automatically benefits
from developments in ASP solving.
%For this approach currently only a direct implementation exists that does not
%support the partial ordering of the reactions.
%
We have shown that our approach is scalable for a certain class of perturbation experiments
and it is already used by the biology research group of Wolfgang Marwan at the
Magdeburg Centre for Systems Biology and
in the context of the GoFORSYS~\citeA{goforsys} project.
It outperforms the direct implementation of the problem while supporting a
broader range of functionality.

In the future we plan to extend our approach in terms of catalysts, as described in~\cite{catalysts}.
% to better adapt to biological needs.
We then want to combine it with the ordering of reactions and also
the automatic addition of species.
Furthermore, we need to improve the approach to be capable of dealing with larger networks,
as all currently tested networks are of small to medium size.
% Another direction of research is the prediction of experiments.
As time series experiments are usually very costly,
we want to investigate how to find optimal experiments to reduce the number of
possible regulatory structures.

%%% Local Variables: 
%%% mode: latex
%%% TeX-master: "paper"
%%% End: 

%% \input{acknowledgments}
\bibliographystyle{acmtrans}
%\bibliography{lit,akku,procs,own}

\begin{thebibliography}{}

\bibitem[\protect\citeauthoryear{Akutsu, Miyano, and Kuhara}{Akutsu
  et~al\mbox{.}}{1999}]{akmiku99a}
{\sc Akutsu, T.}, {\sc Miyano, S.}, {\sc and} {\sc Kuhara, S.} 1999.
\newblock Identification of genetic networks from a small number of gene
  expression patterns under the boolean network model.
\newblock In {\em Pacific Symposium on Biocomputing}. 17--28.

\bibitem[\protect\citeauthoryear{Baral}{Baral}{2003}]{baral02a}
{\sc Baral, C.} 2003.
\newblock {\em Knowledge Representation, Reasoning and Declarative Problem
  Solving}.
\newblock Cambridge University Press.

\bibitem[\protect\citeauthoryear{Baral, Chancellor, Tran, Tran, Joy, and
  Berens}{Baral et~al\mbox{.}}{2004}]{bachtrtrjobe04a}
{\sc Baral, C.}, {\sc Chancellor, K.}, {\sc Tran, N.}, {\sc Tran, N.}, {\sc
  Joy, A.}, {\sc and} {\sc Berens, M.} 2004.
\newblock A knowledge based approach for representing and reasoning about
  signaling networks.
\newblock In {\em Proceedings of the Twelfth International Conference on
  Intelligent Systems for Molecular Biology/Third European Conference on
  Computational Biology (ISMB'04/ECCB'04)}. 15--22.

\bibitem[\protect\citeauthoryear{{Dal Pal{\`u}}, Dovier, and Pontelli}{{Dal
  Pal{\`u}} et~al\mbox{.}}{2009}]{dadopo09a}
{\sc {Dal Pal{\`u}}, A.}, {\sc Dovier, A.}, {\sc and} {\sc Pontelli, E.} 2009.
\newblock Logic programming techniques in protein structure determination:
  Methodologies and results.
\newblock See \citeN{lpnmr09}, 560--566.

\bibitem[\protect\citeauthoryear{Durzinsky, Marwan, and Wagler}{Durzinsky
  et~al\mbox{.}}{2010}]{catalysts}
{\sc Durzinsky, M.}, {\sc Marwan, W.}, {\sc and} {\sc Wagler, A.} 2010.
\newblock Reconstructing extended petri nets.
\newblock {\em Journal of Mathematical Biology\/}~{\em Preprint series: 10-19.
  Available at \texttt{http://www.fma.ovgu.de}}.

\bibitem[\protect\citeauthoryear{Durzinsky, Wagler, and Weismantel}{Durzinsky
  et~al\mbox{.}}{2008}]{DWW_LNCS}
{\sc Durzinsky, M.}, {\sc Wagler, A.}, {\sc and} {\sc Weismantel, R.} 2008.
\newblock A combinatorial approach to reconstruct petri nets from experimental
  data.
\newblock {\em Lecture Notes in Bioinformatics\/}~{\em 5307}, 328 -- 346.

\bibitem[\protect\citeauthoryear{Durzinsky, Wagler, and Weismantel}{Durzinsky
  et~al\mbox{.}}{2010}]{preprint}
{\sc Durzinsky, M.}, {\sc Wagler, A.}, {\sc and} {\sc Weismantel, R.} 2010.
\newblock An algorithmic framework for network reconstruction.
\newblock {\em Theoretical Computer Science\/}.
\newblock In Press, Corrected Proof. Available at
  \texttt{http://www.sciencedirect.com}.

\bibitem[\protect\citeauthoryear{Durzinsky, Wagler, Weismantel, and
  Marwan}{Durzinsky et~al\mbox{.}}{2008}]{duwawema08a}
{\sc Durzinsky, M.}, {\sc Wagler, A.}, {\sc Weismantel, R.}, {\sc and} {\sc
  Marwan, W.} 2008.
\newblock Automatic reconstruction of molecular and genetic networks from
  discrete time series data.
\newblock {\em Biosystems\/}~{\em 93,\/}~3, 181--190.

\bibitem[\protect\citeauthoryear{Dworschak, Grell, Nikiforova, Schaub, and
  Selbig}{Dworschak et~al\mbox{.}}{2008}]{dwgrniscse07a}
{\sc Dworschak, S.}, {\sc Grell, S.}, {\sc Nikiforova, V.}, {\sc Schaub, T.},
  {\sc and} {\sc Selbig, J.} 2008.
\newblock Modeling biological networks by action languages via answer set
  programming.
\newblock {\em Constraints\/}~{\em 13,\/}~1-2, 21--65.

\bibitem[\protect\citeauthoryear{Erdem}{Erdem}{2009}]{erdem09a}
{\sc Erdem, E.} 2009.
\newblock {PHYLO-ASP}: Phylogenetic systematics with answer set programming.
\newblock See \citeN{lpnmr09}, 567--572.

\bibitem[\protect\citeauthoryear{Erdem, Lin, and Schaub}{Erdem
  et~al\mbox{.}}{2009}]{lpnmr09}
{\sc Erdem, E.}, {\sc Lin, F.}, {\sc and} {\sc Schaub, T.}, Eds. 2009.
\newblock {\em Proceedings of the Tenth International Conference on Logic
  Programming and Nonmonotonic Reasoning (LPNMR'09)}. Lecture Notes in
  Artificial Intelligence, vol. 5753. Springer-Verlag.

\bibitem[\protect\citeauthoryear{Erdem and T{\"u}re}{Erdem and
  T{\"u}re}{2008}]{erdtur08a}
{\sc Erdem, E.} {\sc and} {\sc T{\"u}re, F.} 2008.
\newblock Efficient haplotype inference with answer set programming.
\newblock In {\em Proceedings of the Twenty-third National Conference on
  Artificial Intelligence (AAAI'08)}, {D.~Fox} {and} {C.~Gomes}, Eds. AAAI
  Press, 436--441.

\bibitem[\protect\citeauthoryear{Gebser, Guziolowski, Ivanchev, Schaub, Siegel,
  Thiele, and Veber}{Gebser et~al\mbox{.}}{2010}]{geguivscsithve10a}
{\sc Gebser, M.}, {\sc Guziolowski, C.}, {\sc Ivanchev, M.}, {\sc Schaub, T.},
  {\sc Siegel, A.}, {\sc Thiele, S.}, {\sc and} {\sc Veber, P.} 2010.
\newblock Repair and prediction (under inconsistency) in large biological
  networks with answer set programming.
\newblock In {\em Proceedings of the Twelfth International Conference on
  Principles of Knowledge Representation and Reasoning (KR'10)}, {F.~Lin} {and}
  {U.~Sattler}, Eds. AAAI Press, 497--507.

\bibitem[\protect\citeauthoryear{Gebser, Kaminski, Kaufmann, Ostrowski, Schaub,
  and Thiele}{Gebser et~al\mbox{.}}{}]{potasscoManual}
{\sc Gebser, M.}, {\sc Kaminski, R.}, {\sc Kaufmann, B.}, {\sc Ostrowski, M.},
  {\sc Schaub, T.}, {\sc and} {\sc Thiele, S.}
\newblock A user's guide to \texttt{gringo}, \texttt{clasp}, \texttt{clingo},
  and \texttt{iclingo}.
\newblock Available at {\texttt{http://potassco.sourceforge.net}}.

\bibitem[\protect\citeauthoryear{Gebser, Kaufmann, and Schaub}{Gebser
  et~al\mbox{.}}{2009}]{gekasc09a}
{\sc Gebser, M.}, {\sc Kaufmann, B.}, {\sc and} {\sc Schaub, T.} 2009.
\newblock Solution enumeration for projected {B}oolean search problems.
\newblock In {\em Proceedings of the Sixth International Conference on
  Integration of AI and OR Techniques in Constraint Programming for
  Combinatorial Optimization Problems (CPAIOR'09)}, {W.~{van Hoeve}} {and}
  {J.~Hooker}, Eds. Lecture Notes in Computer Science, vol. 5547.
  Springer-Verlag, 71--86.

\bibitem[\protect\citeauthoryear{Gebser, Schaub, Thiele, Usadel, and
  Veber}{Gebser et~al\mbox{.}}{2008}]{gescthusve08b}
{\sc Gebser, M.}, {\sc Schaub, T.}, {\sc Thiele, S.}, {\sc Usadel, B.}, {\sc
  and} {\sc Veber, P.} 2008.
\newblock Detecting inconsistencies in large biological networks with answer
  set programming.
\newblock In {\em Proceedings of the Twenty-fourth International Conference on
  Logic Programming (ICLP'08)}, {M.~{Garcia de la Banda}} {and} {E.~Pontelli},
  Eds. Lecture Notes in Computer Science, vol. 5366. Springer-Verlag, 130--144.

\bibitem[\protect\citeauthoryear{Gelfond}{Gelfond}{2008}]{gelfond08a}
{\sc Gelfond, M.} 2008.
\newblock Answer sets.
\newblock In {\em Handbook of Knowledge Representation}, {V.~Lifschitz},
  {F.~{van Hermelen}}, {and} {B.~Porter}, Eds. Elsevier, Chapter~7, 285--316.

\bibitem[\protect\citeauthoryear{Gifford and Jaakkola}{Gifford and
  Jaakkola}{2001}]{gifjaa01a}
{\sc Gifford, D.} {\sc and} {\sc Jaakkola, T.} 2001.
\newblock Using graphical models and genomic expression data to statistically
  validate models of genetic regulatory networks.
\newblock In {\em Pacific Symposium on Biocomputing}. 422--433.

\bibitem[\protect\citeauthoryear{??}{goforsys}{}]{goforsys}
goforsys.
\newblock \texttt{http://www.goforsys.de}.

\bibitem[\protect\citeauthoryear{Laubenbacher and Stigler}{Laubenbacher and
  Stigler}{2004}]{lausti04a}
{\sc Laubenbacher, R.} {\sc and} {\sc Stigler, B.} 2004.
\newblock A computational algebra approach to the reverse engineering of gene
  regulatory networks.
\newblock {\em Journal of Theoretical Biology\/}~{\em 229,\/}~4, 523 -- 537.

\bibitem[\protect\citeauthoryear{Liang, Fuhrman, and Somogyi}{Liang
  et~al\mbox{.}}{1998}]{lifuso98a}
{\sc Liang, S.}, {\sc Fuhrman, S.}, {\sc and} {\sc Somogyi, R.} 1998.
\newblock Reveal, a general reverse engineering algorithm for inference of
  genetic network architectures.
\newblock {\em Pacific Symposium on Biocomputing\/}~{\em 3}, 18--29.

\bibitem[\protect\citeauthoryear{Lin and Zhao}{Lin and Zhao}{2004}]{linzha04a}
{\sc Lin, F.} {\sc and} {\sc Zhao, Y.} 2004.
\newblock {ASSAT}: computing answer sets of a logic program by {SAT} solvers.
\newblock {\em Artificial Intelligence\/}~{\em 157,\/}~1-2, 115--137.

\bibitem[\protect\citeauthoryear{Marwan, Wagler, and Weismantel}{Marwan
  et~al\mbox{.}}{2008}]{marwagwei08a}
{\sc Marwan, W.}, {\sc Wagler, A.}, {\sc and} {\sc Weismantel, R.} 2008.
\newblock A mathematical approach to solve the network reconstruction problem.
\newblock {\em Mathematical Methods of Operatios Research\/}~{\em 67,\/}~1,
  117--132.

\bibitem[\protect\citeauthoryear{Ostrowski}{Ostrowski}{2011}]{website}
{\sc Ostrowski, M.} 2011.
\newblock Asp encoding for automatic network reconstruction.
\newblock \texttt{http://www.cs.uni-potsdam.de/wv/NetworkReconstruction/}.

\bibitem[\protect\citeauthoryear{Pe’er, Regev, Elidan, and Friedman}{Pe’er
  et~al\mbox{.}}{2001}]{pereelfr01a}
{\sc Pe’er, D.}, {\sc Regev, A.}, {\sc Elidan, G.}, {\sc and} {\sc Friedman,
  N.} 2001.
\newblock Inferring subnetworks from perturbed expression profiles.
\newblock {\em Bioinformatics\/}~{\em 17,\/}~suppl 1, 215--224.

\bibitem[\protect\citeauthoryear{Repsilber, Liljenström, and
  Andersson}{Repsilber et~al\mbox{.}}{2002}]{relian02a}
{\sc Repsilber, D.}, {\sc Liljenström, H.}, {\sc and} {\sc Andersson, S.}
  2002.
\newblock Reverse engineering of regulatory networks: simulation studies on a
  genetic algorithm approach for ranking hypotheses.
\newblock {\em Biosystems\/}~{\em 66,\/}~1-2, 31 -- 41.

\bibitem[\protect\citeauthoryear{Schaub and Thiele}{Schaub and
  Thiele}{2009}]{schthi09a}
{\sc Schaub, T.} {\sc and} {\sc Thiele, S.} 2009.
\newblock Metabolic network expansion with {ASP}.
\newblock In {\em Proceedings of the Twenty-fifth International Conference on
  Logic Programming (ICLP'09)}, {P.~Hill} {and} {D.~Warren}, Eds. Lecture Notes
  in Computer Science, vol. 5649. Springer-Verlag, 312--326.

\bibitem[\protect\citeauthoryear{Simons, Niemel{\"a}, and Soininen}{Simons
  et~al\mbox{.}}{2002}]{siniso02a}
{\sc Simons, P.}, {\sc Niemel{\"a}, I.}, {\sc and} {\sc Soininen, T.} 2002.
\newblock Extending and implementing the stable model semantics.
\newblock {\em Artificial Intelligence\/}~{\em 138,\/}~1-2, 181--234.

\bibitem[\protect\citeauthoryear{Syrj{\"a}nen}{Syrj{\"a}nen}{}]{lparseManual}
{\sc Syrj{\"a}nen, T.}
\newblock Lparse 1.0 user's manual.
\newblock http://www.tcs.hut.fi/Software/smodels/lparse.ps.gz.

\bibitem[\protect\citeauthoryear{Wagler}{Wagler}{2011}]{wagler11a}
{\sc Wagler, A.} 2011.
\newblock Prediction of network structure.
\newblock {\em Modeling in Systems Biology\/}~{\em 16}, 307--336.

\bibitem[\protect\citeauthoryear{Yeung, Tegnér, and Collins}{Yeung
  et~al\mbox{.}}{2002}]{yeteco02a}
{\sc Yeung, M.}, {\sc Tegnér, J.}, {\sc and} {\sc Collins, J.} 2002.
\newblock Reverse engineering gene networks using singular value decomposition
  and robust regression.
\newblock {\em Proceedings of the National Academy of Sciences of the United
  States of America\/}~{\em 99,\/}~9, 6163--6168.

\end{thebibliography}

%\newpage
%\input{appendix}

\end{document}